\documentclass[10pt,twocolumn,letterpaper]{article}

\usepackage[pagenumbers]{cvpr} 
\usepackage{graphicx}
\usepackage{amsmath}
\usepackage{soul}
\usepackage{command}
\usepackage{comment}
\usepackage{amssymb}
\usepackage{booktabs}
\usepackage{multirow}
\usepackage{bm}

\usepackage[ruled, vlined, linesnumbered]{algorithm2e}
\SetKwComment{Comment}{$\triangleright$\ }{}

\usepackage[pagebackref,breaklinks,colorlinks]{hyperref}
\newcommand{\name}[0]{SD\xspace}
\usepackage{soul,color}
\usepackage[capitalize]{cleveref}
\crefname{section}{Sec.}{Secs.}
\Crefname{section}{Section}{Sections}
\Crefname{table}{Table}{Tables}
\crefname{table}{Tab.}{Tabs.}

\newtheorem{assumption}{Assumption}

\def\cvprPaperID{8143} 
\def\confName{CVPR}
\def\confYear{2023}

\begin{document}

\title{Improving Adversarial Transferability with Scheduled Step Size and Dual Example}

\author{Zeliang Zhang\\
University of Rochester\\
\and
Peihan Liu\\
Harvard University\\
\and
Xiaosen Wang\thanks{Corresponding author}\\
HUST\\
\and
Chenliang Xu\textsuperscript{$\star$}\\
University of Rochester\\
}


\maketitle


\begin{abstract}
   Deep neural networks are widely known to be vulnerable to adversarial examples, especially showing significantly poor performance on adversarial examples generated under the white-box setting. However, all white-box attack methods rely heavily on the target model and quickly get stuck in local optima, resulting in poor adversarial transferability. The momentum-based methods and their variants are proposed to escape the local optima for better transferability. In this work, we notice that the transferability of adversarial examples generated by iterative fast gradient sign method (I-FGSM) exhibits a decreasing trend when increasing the number of iterations. Motivated by this finding, we argue that the information of adversarial perturbations near the benign sample, especially the direction,  benefit more on the transferability.  Thus, we propose a novel strategy, which uses the \textbf{S}cheduled step size and the \textbf{D}ual example (SD), to fully utilize the adversarial information near the benign sample.  Our proposed strategy can be easily integrated with existing adversarial attack methods for better adversarial transferability. Empirical evaluations on the standard ImageNet dataset demonstrate that our proposed method can significantly enhance the transferability of existing adversarial attacks.
\end{abstract}

\section{Introduction}
Adversarial examples~\cite{szegedy2014intriguing,goodfellow2014explaining}, crafted by adding human-imperceptible perturbations on benign samples to fool deep neural networks (DNNs), have drawn more research interest in recent years ~\cite{agarwal2022exploring,jang2022strengthening,zou2022making}. On the one hand, the existence of adversarial examples identifies the vulnerability of DNNs, throwing out the severe security issues to be solved urgently on many areas, including the autonomous driving~\cite{im2022adversarial}, facial authentication~\cite{joos2022adversarial}, object detection~\cite{Hu_2021_ICCV}, \etal. The adversarial examples also generalized to other models ~\cite{papernot2017practical,xie2019improving,wu2021improving}, where the adversarial examples crafted by one model can also fool other models. On the other hand, the adversarial examples help evaluate the robustness of DNNs and enhance the robustness of DNNs through adversarial training ~\cite{shafahi2019adversarial,wong2020fast}. Great research interest has arisen in how to generate adversarial examples with high transferability for a better understanding of the DNNs~\cite{wu2020towards}.  

According to the knowledge about the victim model, the existing adversarial attacks can be categorized into two settings, namely the white-box~\cite{goodfellow2014explaining,moosavi2016deepfool,zhang2019defending} and the black-box settings~\cite{ilyas2018black,rahmati2020geoda,shi2019curls}. With the full knowledge of the victim model, including the architecture, the parameter, the training loss function, \etc, most of the existing adversarial attacks under the white-box setting usually exhibit a great performance but behave poorly facing the models without knowing any information under the black-box setting~\cite{wang2021admix, lin2019nesterov}. The low adversarial transferability makes applying such existing adversarial attacks inefficient in real-world applications, where the victim models are usually unknown to attackers~\cite{naseer2021improving}.

Some methods have been proposed to enhance the transferability of adversarial examples, such as gradient-based attacks with momentum~\cite{wang2021enhancing,han2022sampling}, input transformation-based attacks~\cite{peng2022speckle,wang2022pre}, ensemble attacks~\cite{shen2021bbas,che2020new}, \etc. Among them, the gradient-based methods mainly utilize the momentum (MI-FGSM)~\cite{dong2018boosting} or pre-computation (NI-FGSM)~\cite{Lin2020Nesterov} to help escape the local optima for better adversarial transferability. The input transformation~\cite{xie2019feature,dong2019evading} methods analogize the crafting of adversarial examples to the generalization of DNNs, thus diversifying  the inputs to generate the adversarial examples with high transferability. 

Different from the above adversarial attacks, we study adversarial transferability from a different perspective. From a simple experiment, we observe that the transferability of adversarial examples crafted using vanilla iterative adversarial attacks demonstrate a trend from high to low, which indicates that the directions of the adversarial perturbations near the benign sample are more transferable among different models. Motivated by this,  we propose a novel strategy with the \textbf{S}cheduled step size and \textbf{D}ual example (SD), which fully explores and utilizes the adversarial perturbations with high transferability near the benign sample. Also, it can be easily integrated on the existing adversarial attack methods to enhance the performance on transferability.   
  
We apply our proposed method to several generally used adversarial attack methods and conduct proof experiments on the standard ImageNet dataset. The results demonstrate that our proposed method can remarkably boost the existing adversarial attack methods on transferability.

\section{Related Work}
In this section, we provide a brief overview of adversarial attack and defense.
\subsection{Adversarial Attack}
Since Szegedy~\etal~\cite{szegedy2014intriguing} identified the vulnerability of DNNs to adversarial examples, numerous adversarial attack methods have been proposed, including white-box attack~\cite{goodfellow2014explaining,moosavi2016deepfool,kurakin2017adversarial, madry2017towards,croce2020reliable}, transfer-based attack~\cite{dong2018boosting, xie2019improving,dong2019evading,wang2021enhancing,wang2021admix,wang2021boosting}, score-based attack~\cite{ilyas2018black,uesato2018adversarial,li2019nattack,cheng2019improving}, decision-based attack~\cite{brendel2017decision,cheng2019query,li2020qeba,wang2022triangle,maho2021surfree}, \etc. Among the above attacks, a transfer-based attack, as a black-box attack, does not need access to the target model, making it popular to attack the deep models in the real world. We provide an overview of the transfer-based attacks as follows.
 
\textbf{Gradient-based Attacks.} 
Fast Gradient Sign Method (FGSM)~\cite{goodfellow2014explaining}, which is the first gradient-based attack, crafts adversarial examples by adding a large perturbation in the gradient direction to the benign sample. To improve its white-box attack performance, it is further extended to an iterative Version (I-FGSM)~\cite{kurakin2017adversarial} but shows poor transferability.  With the potential application of transfer-based attacks in the real world, various methods have been proposed to improve the transferability of adversarial examples. Dong~\etal~\cite{dong2018boosting} introduce the momentum into I-FGSM, dubbed Momentum I-FGSM (MI-FGSM), to stabilize the optimization and escape the local optima. Lin~\etal~\cite{Lin2020Nesterov} utilize
Nesterov accelerated gradient to look ahead, denoted as NI-FGSM, which exhibits better transferability. Wang~\etal~\cite{wang2021enhancing} propose the definition of gradient variance, which could be used to tune the gradient for MI-FGSM (VMI-FGSM) and NI-FGSM (VNI-FGSM). Wang~\etal~\cite{wang2021boosting} enhance the momentum with the gradient of multiple samples in the neighborhood of the current adversarial example. 

\textbf{Input transformation-based attacks.} Inspired by the data augmentation in training, various input transformation-based attacks are proposed to improve the transferability, which can be combined with the above gradient-based attacks. For instance, the diverse input method (DIM)~\cite{xie2019improving} resizes the input image into a random size, which will be padded to a fixed size for gradient calculation. Translation invariant method (TIM)~\cite{dong2019evading} adopts Gaussian smooth on the gradient to approximate the average gradient of a set of translated images to update the adversary. Scale-invariant method (SIM)~\cite{Lin2020Nesterov} calculates the gradient on a set of scaled images. \textit{Admix}~\cite{wang2021admix} adds a small portion of images from other categories to the input image to obtain several images for gradient calculation. 

\textbf{Ensemble attacks.} Liu~\etal~\cite{liu2016delving} first find that ensemble attack, which generates adversarial examples on multiple models, can lead to better transferability. Xiong~\etal~\cite{xiong2022stochastic} reduces the gradient variance among various models to boost ensemble attack. Long~\etal~\cite{long2022frequency} augment the model in the frequency domain to obtain more diverse substitute models.

\textbf{Others.}
Gao~\etal~\cite{gao2020patch} reuses the cut noise introduced by clip operation. Wu~\etal~\cite{wu2020boosting} argues that the attack should maximize the difference of the attention map between benign samples and adversarial examples. 

\subsection{Adversarial Defense}
On the other hand, to mitigate the threat of adversarial attacks, a variety of defense methods have been proposed, \eg adversarial training~\cite{goodfellow2014explaining,madry2017towards,zhang2019theoretically,wang2019improving}, input pre-processing~\cite{guo2017countering}, certified defense~\cite{cohen2019certified} \etc.  Adversarial training has been shown as one of the most effective methods to improve adversarial robustness~\cite{athalye2018obfuscated,croce2020reliable}. It is first proposed by Madry \etal \cite{madry2017towards} that including the adversarial examples in the training data enhances the adversarial robustness.  Wong \etal \cite{wong2020fast} use the fast gradient sign method with random initialization to generate adversarial examples for efficient adversarial training.

\section{Methods}
In this section, we introduce our motivation and provide a detailed description of the proposed strategy \name. 

\subsection{Motivation}
\label{sec:motivation}

Given a target deep model $f$ and a benign sample $x$ with the ground-truth label $y$, adversarial attacks find an example $x^{adv}$ similar to $x$ (\ie, $x^{adv} \in \mathcal{B}_\epsilon(x)=\{x' : \|x'-x\|_p\leq \epsilon\}$ where $\|\cdot\|_p$ refers to the $p-$norm) such that the model gives different predictions, \ie, $f(x)=y\ne f(x^{adv})$.  Gradient-based adversarial attacks (\eg, FGSM~\cite{goodfellow2014explaining}, I-FGSM~\cite{kurakin2017adversarial}, MI-FGSM~\cite{dong2018boosting}, NI-FGSM~\cite{Lin2020Nesterov}, \etc) can generate the adversarial examples $x^{adv}$ efficiently. Goodfellow \etal ~\cite{goodfellow2014explaining} first propose the gradient sign method (FGSM) to attack the deep learning model efficiently. However, single-step optimization for adversarial examples may not find the optima, which fails to fool the models under the white-box setting, within the box constraint $\mathcal{B}_\epsilon(x)$. Iterative fast gradient sign method (I-FGSM) employs the iterative process to optimize $x^{adv}$ and can achieve a attack success rate of $100\%$ or nearly $100\%$.  

Suppose $\mathcal{L}(\cdot,\cdot)$ is the classification loss, I-FGSM adopts the iterative optimization process to craft an adversarial example $x^{adv}$ which maximizes $\mathcal{L}(f(x^{adv}),y)$ as follows,
%
\begin{equation}
\label{eq:ifgsm}
    x^{adv}_{t+1}=\Pi_{\mathcal{B}_\epsilon(x)}\left\{x^{adv}_{t}+\alpha \cdot \operatorname{sign}\left(\nabla_x\mathcal{L}\left(f\left(x^{adv}_{t}\right),y\right)\right)\right\}, 
\end{equation}
where $\Pi$ is the projector function, $\alpha$ is the step size, and $x_t^{adv}$ is the adversarial example at the $t$-th iteration. 

\begin{figure}[tbp]
    \centering
    \includegraphics[scale=0.5]{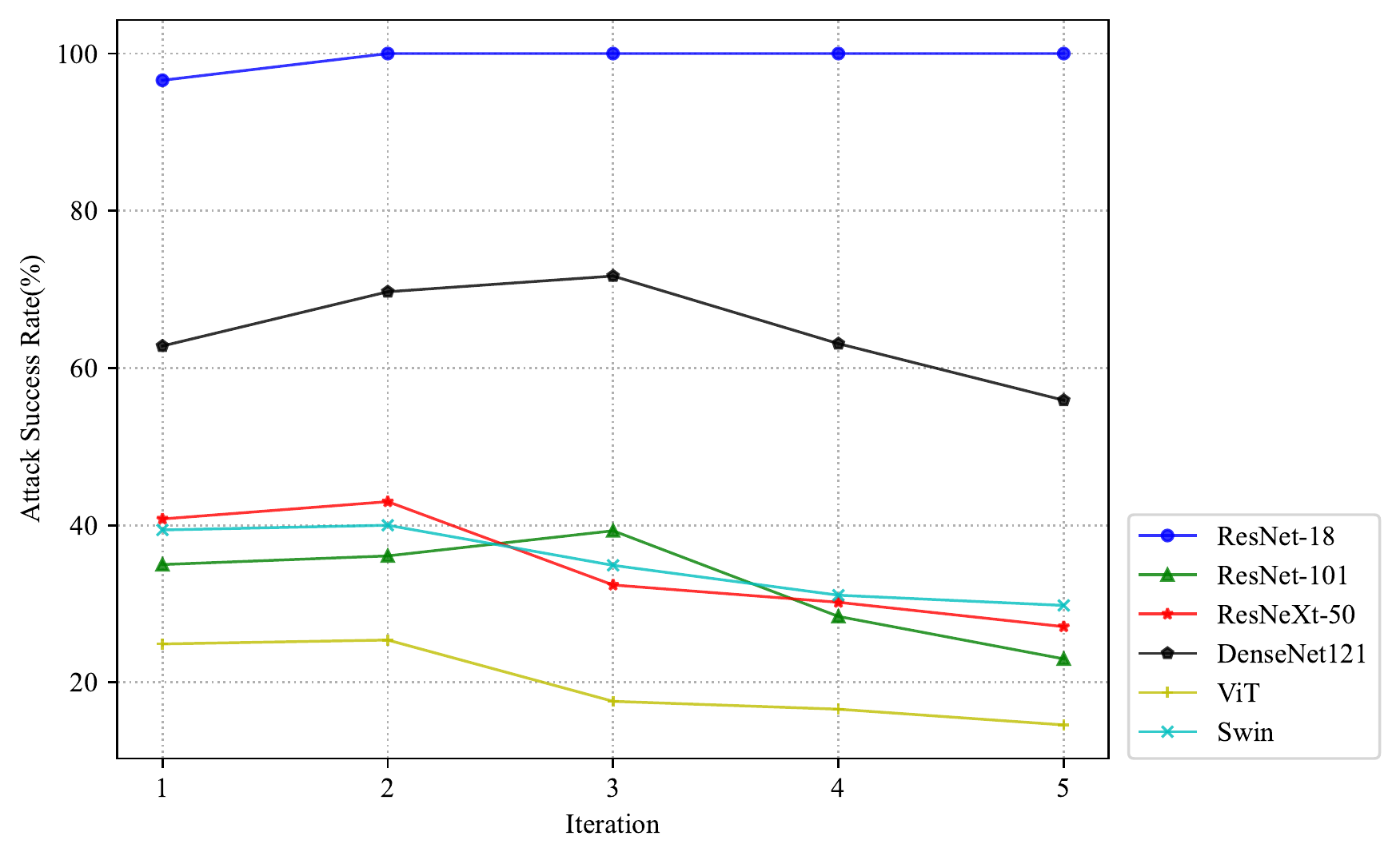}
    \caption{Varying the maximum number of iteration, the attack success rate of six models on the  adversarial examples generated by ResNet-18.}
    \label{fig:motivation}
\end{figure}

To further explore how the number of iterations affects the attack performance, we conduct I-FGSM attack on ImageNet dataset with different iterations ($T=1 \to 5$ iterations). The adversarial examples are generated on ResNet-18~\cite{he2016deep}, which are used to attack other models (ResNet-101~\cite{he2016deep}, ResNext-50~\cite{xie2017aggregated}, DenseNet-121~\cite{huang2017densely}, ViT~\cite{dosoViTskiy2020image}, Swin~\cite{liu2021swin}) with the perturbation budget $\epsilon=16/255$ and step size $\alpha=\epsilon/T$. As shown in Fig.~\ref{fig:motivation}, I-FGSM  achieves $100\%$ attack success rate on the white-box model (ResNet-18) with 2 iterations. Also, it is obvious that the transferabilty of adversarial examples generated by FGSM, \ie, $T=1$, are better than that of I-FGSM with multiple steps ($T \ge 4$). Since the difference of updates on $x^{adv}$ in each iteration  only falls on the direction, \ie, the sign of the gradient, it naturally inspires us the following assumption,  

\begin{assumption}
    The direction of the adversarial perturbation near the benign sample benefits more on adversarial transferability than that far away.
    \label{ass:density}
\end{assumption}



Lin \etal \cite{Lin2020Nesterov} analogize the search for adversarial examples with the training process of models and the transferability of adversarial examples with the generalization ability. Training on the same dataset, \eg, ImageNet dataset, different models usually have similar attentions on the features. However, the adversarial example is one category of the out-of-distribution (OOD) data. Evaluated on such OOD datasets, different models may have different performances, \ie, different attentions on the features.  With an increasing number of iterations, the distribution shifts become large and the performance difference, \ie, attention on the features, will also become large. Thus, the adversarial perturbation added on examples computed by the white-box model, that are far away from the benign sample, are not generally easy to transfer to be efficient for other models. 

A natural method to alleviate the above problem, caused by  distribution shifts, is to use  decreasing step sizes to fully utilize the adversarial perturbation direction near the benign sample. Thus, we use several kinds of decreasing sequence as the step sizes $\alpha$  to optimize $x^{adv}$ in (\ref{eq:ifgsm})  and evaluate the attack performances on aforementioned models. 

\begin{figure}[tbp]
    \centering
    \includegraphics[scale=0.5]{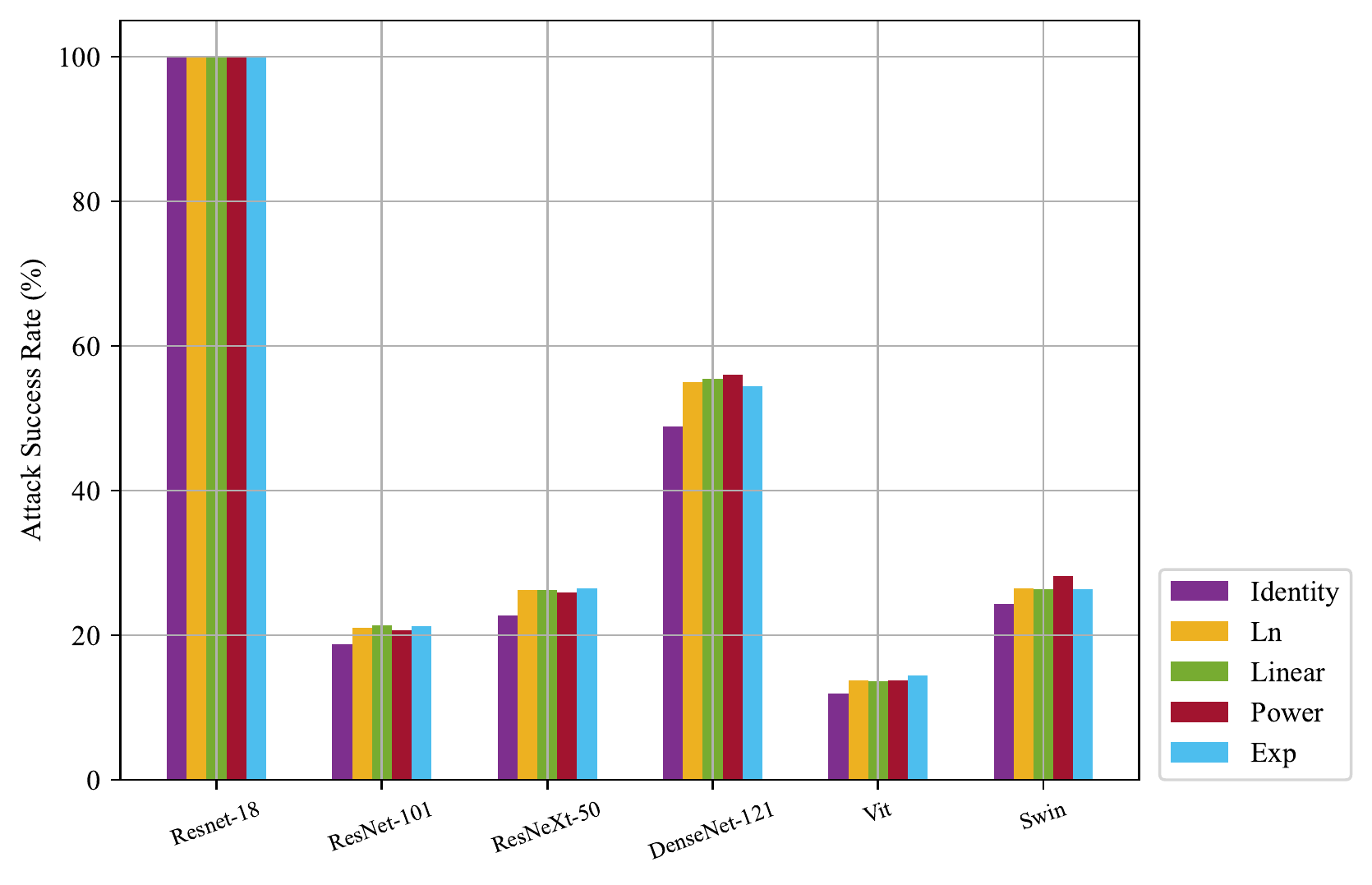}
    \caption{With different decreasing scheduled step sizes, the attack success rate of six model on the adversarial examples generated by ResNet-18.}
    \label{fig:decreasing}
\end{figure}

We adopt four decreasing sequences: logarithm ($\alpha_i=\ln(T-i)$), linear ($\alpha_i=T-i$), power ($\alpha_i=(T-i)^2$), and exponential ($\alpha_i=e^{T-i}$) sequences. We normalize each sequence by dividing the sum to satisfy $\mathcal{B}_\epsilon(x)$.  As shown in results in Fig. \ref{fig:decreasing},  the scheduled decreasing step size improves the adversarial attack success rates under the black-box setting, with an average of $2.66\%$ for (Ln), $2.75\%$ for (Linear), $2.98\%$ for (Power) and $2.72\%$ for (Exp). However, the results are still worse than that of FGSM and I-FGSM with a few iterations on the adversarial transferability. The reason why the proposed method does not work as expected is that, there is a huge gap between the magnitude of the first step  and the magnitude of the last step,
which leads to insufficient updates on the last several iterations of I-FGSM. The huge gap of the schedule step sizes between the early and last stages are also discussed in our ablation study in Section 4.6.  So, besides employing the scheduled decreasing steps, is there any alternative way to fully utilize the information near the benign sample to enhance the adversarial transferability?

\begin{figure*}[htbp]
    \centering
    \includegraphics[width=\textwidth]{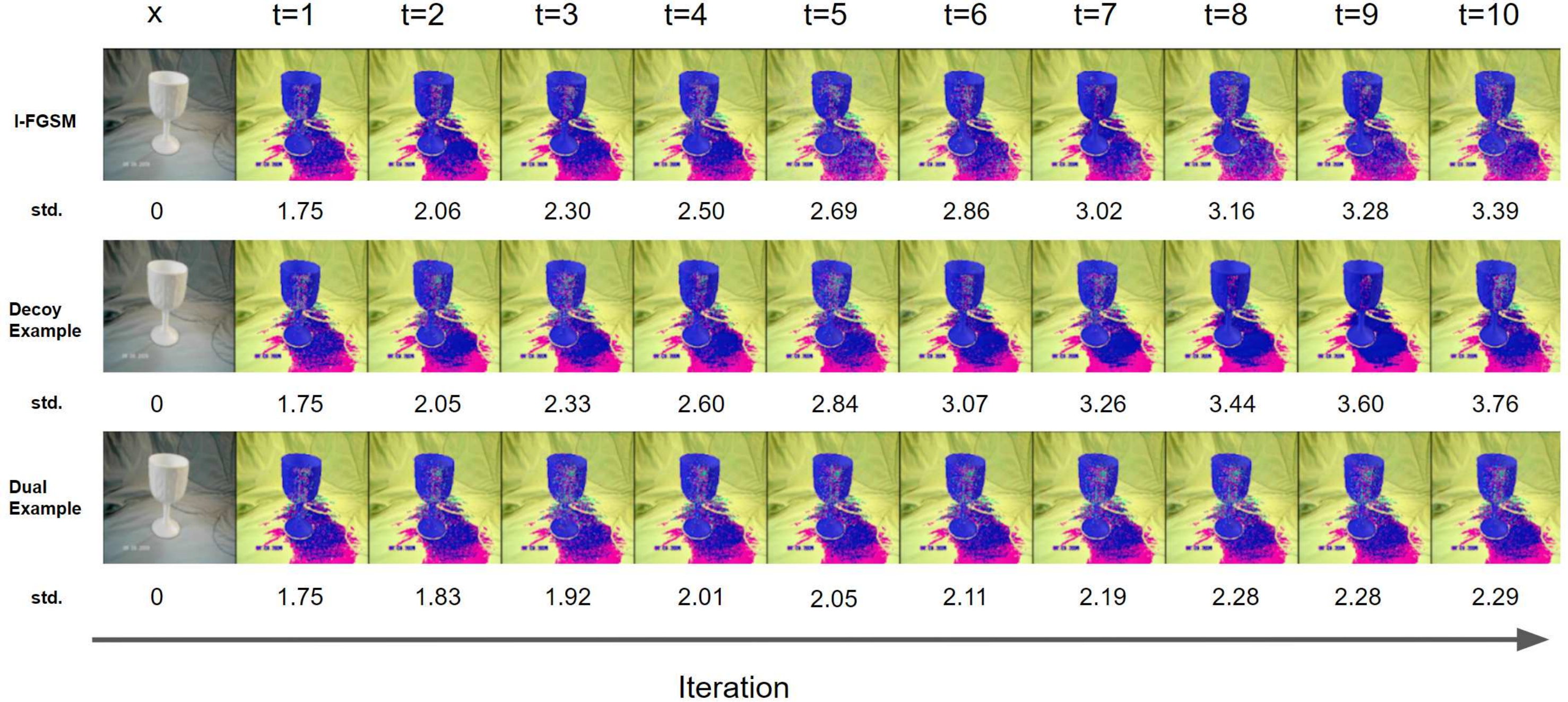}
    \caption{Visualization of the adversarial examples with perturbations for each iteration. The strength of the color on the mask represents the adversarial perturbation. The sharp colors, especially red, indicate the large perturbations, and the soft colors, \eg, yellow, indicate the small perturbations. \textbf{Please view on a color screen.}}
    \label{fig:example_iteration}
\end{figure*}

\subsection{Method}
\label{sec:method}
To fully utilize the information, \ie, the direction of adversarial perturbations with high transferability, except for using the decreasing step size, we can also use the scheduled increasing step size. 

\noindent \textbf{Scheduled  step size}: As aforementioned, we hope to utilize the scheduled decreasing step size to rearrange larger weights on the early updates for $x^{adv}$, where the direction of the updates, \ie, the adversarial perturbation, is more transferable. Moreover, we want the updates on the later stages to sufficiently useful. It turns out that using the scheduled increasing step sizes can achieve the two goals simultaneously. More specifically, we sample more directions of the adversarial perturbation with high transferability and put them on use in the later iterations to enhance the transferability of adversarial examples. 

In particular, rather than using an identity step size in 
 (\ref{eq:ifgsm}), we made the step sizes as a increasing sequence to find $x^{adv}$ as follows,
\begin{equation}
\label{eq:adaptive_attack}
    x^{adv}_{t+1}=Clip_{\epsilon}\left\{x^{adv}_{t}+\hat{\alpha}_{t} sign\left(\nabla_{x^{adv}_{t}}\mathcal{L}\left(f\left(x^{adv}_{t}\right),y\right)\right)\right\},
\end{equation}
where $\hat{\alpha}$ is the scheduled  step size sequence.  

As shown in Fig. \ref{fig:decreasing}, due to the huge gap of the step size between the early and last several iterations, the inefficient updates on the adversarial example on the last several iterations do not gain a lot for the transferability. However, with the increasing sequence as the scheduled step size, the updates on the last few iterations becomes sufficient. Also, due to the small step size on the first few iterations, $x^{adv}$ stays near around $x$, meaning that we can sample more directions of adversarial perturbations with high transferability.  Then, the next question is: how can we utilize the sampled adversarial perturbations with high transferability?


\noindent \textbf{Dual example}: In order to make full use of the transferable directions of adversarial perturbations, we propose a novel strategy, namely dual example, which uses two examples in parallel but applies different  updates on them. 

Specifically, we initialize two examples $x^{adv}=x^{dual}=x$, namely the decoy and dual example, respectively, in our method. In each iteration,  we use the decoy example, \ie, $x^{adv}$, to explore and sample adversarial perturbations, and use $x^{dual}$ to fully utilize them. Due to the scheduled increasing step size, the early sampled adversarial perturbations using small step size near the benign sample are  usually more transferable than the later ones. Thus, we average all the previous adversarial perturbations as an ensemble result to compute the update on the current iteration. There are two advantages using this way. On the one hand, even for the last few steps, where $x^{adv}$ is quite far away from $x$ and the adversarial perturbations usually behave poorly on the transferability, the actual updates on $x^{dual}$ will not be affected much from the ensemble result on the early updates.   On the other hand, the large step size, for the later stage, will help $x^{dual}$ jump over the local optima as well as making each update more sufficiently useful. It addresses the major drawback using the scheduled decreasing step size as introduced in the previous section.    


From the above analysis and design, we present our algorithm in Alg.~\ref{alg:alg}. We  use the same scheduled step size to update the dual examples to make the optimization of $x^{adv}$ coordinated with $x$. 

Further, we visualize the benign sample and adversarial examples in each iterations and highlight the adversarial perturbations in Fig.~\ref{fig:example_iteration}. The sharp colors, including the light green, pink and red, represent large perturbations,  while the soft colors fulfilled at the background, especially the yellow, represent small perturbations. Besides, we also compute the standard deviation (std) of the previous perturbations for the current iteration.  From the figure (first two rows of images), it can be observed that the attackers' feature attention varies a lot between different iterations. For instance, there are large adversarial perturbations on the \textit{table} of the image in the $5,6,8,9,10$-th iteration. For this,  we argue that the large variance on perturbations between different iterations comes from the data distribution shifts, \ie, from natural examples to adversarial examples. The data distribution shifts lead to unstable performance on the feature attention for models, which also causes the drop of performance on the adversarial transferablity. On the other hand, the proposed \name strategy, \ie, images of the last row in Fig.~\ref{fig:example_iteration}, can alleviate this problem. It can be clearly observed that the variance does not change significantly, with stable feature attention. 

\begin{algorithm}[t] 
\SetKwInput{KwInput}{Input}                
\SetKwInput{KwOutput}{Output}              
    \caption{I-FGSM with Scheduled Adaptive Step Size and Dual Example}
	\label{alg:alg}
    \LinesNumbered 
    \KwInput{Classifier $\bm{f}(\cdot)$; The benign sample $x$ with ground-truth label $y$; Loss function $\bm{L}(\cdot,\cdot)$; The number of iterations $T$; Decreasing sequence $\hat{\alpha}$; }
	\KwOutput{An transferable adversarial example.}  
	initialize a random perturbation $\delta_{0}$, dual sample $x^{dual}_{0}=x_{0}=x$, moving average update $m^{dual}_{0}=m_0=0$, step $t=0$;\\
    \While{$t<T$}{
        $g=\frac{\partial \bm{L}(\bm{f}(x_{t}), y)}{\partial x_{t}}$\Comment*[r]{explore and sample features}
        $m^{dual}_{t+1}=\frac{m^{dual}_{t}t+g}{t+1}$\Comment*[r]{exploit and smooth the features}
        $x_{t+1}=x_{t}+\hat{\alpha}_{t}\bm{sign}(g)$\Comment*[r]{ update input sample}
       $x^{dual}_{t+1}=x^{dual}_{t}+\hat{\alpha}_{t}\bm{sign}(m^{dual}_{t+1})$\Comment*[r]{ update the dual sample }
       $t \gets t+1$;
    }
    \KwRet{$x_{T}^{dual}$.}
\end{algorithm}

\subsection{Relationship to Existing Methods}
The mechanism of scheduled adaptive step size and dual example can be easily integrated with existing methods such as other gradient-based methods, including MI-FGSM and NI-FGSM, and transformation-based methods, including TIM, DIM, and SIM.

\noindent \textbf{Gradient-based methods}: A notable difference between I-FGSM and the others is that MI-FGSM and NI-FGSM replace the gradients with momentums to enhance the performance. To integrate our method with them, it only needs to change the line $4$ in Alg. \ref{alg:alg} with the adaptive momentum as follows,
\begin{equation}
    \label{eq:adaptive_momentum}
    m_{t+1} = \mu m_{t}+g.
\end{equation}

\noindent \textbf{Transformation-based methods}: The transformation-based methods use $\mathcal{T}(x)$ instead of $x$ to compute $g$, where $\mathcal{T}(\cdot)$ is an input transformation function. It only needs to change $x_{t}$ of the line $3$ in Alg. \ref{alg:alg} with $\mathcal{T}(x_{t})$.

\noindent \textbf{Ensemble attack methods}: The ensemble attacks generate adversarial examples by attacking several models in parallel. To integrate our method in the ensemble attack, we can change the computation of gradient in line $3$ from the single-model setting to the ensemble-model setting, \ie $g=\frac{\partial \bm{L}(\frac{1}{N}\sum\limits_{n=1}^{N}\bm{f}_{i}(x_{t}), y)}{\partial x_{t}}$, where $N$ indicates the number of ensemble models.


\section{Experiments}

\begin{figure*}[htbp]
    \centering
    \includegraphics[width=\textwidth]{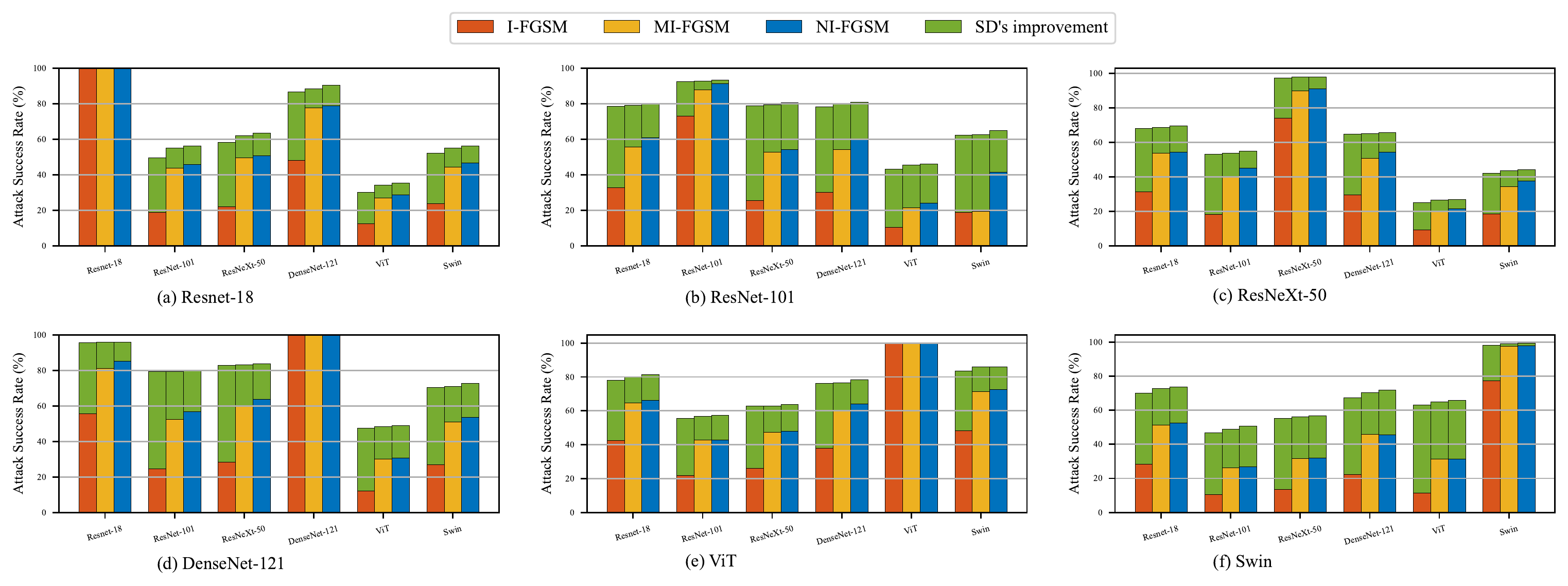}
    \caption{The Attack success rate (\%) of baseline models
on the crafted adversarial examples generated by I-FGSM, MI-FGSM, NI-FGSM and our proposed method under single model setting. We generate the adversarial examples on one model and evaluate on the other five models. \textbf{Please view on a color screen.}}
    \label{fig:baseline_performance}
    \end{figure*} 
\begin{figure*}[htbp]
    \centering
    \includegraphics[width=\textwidth]{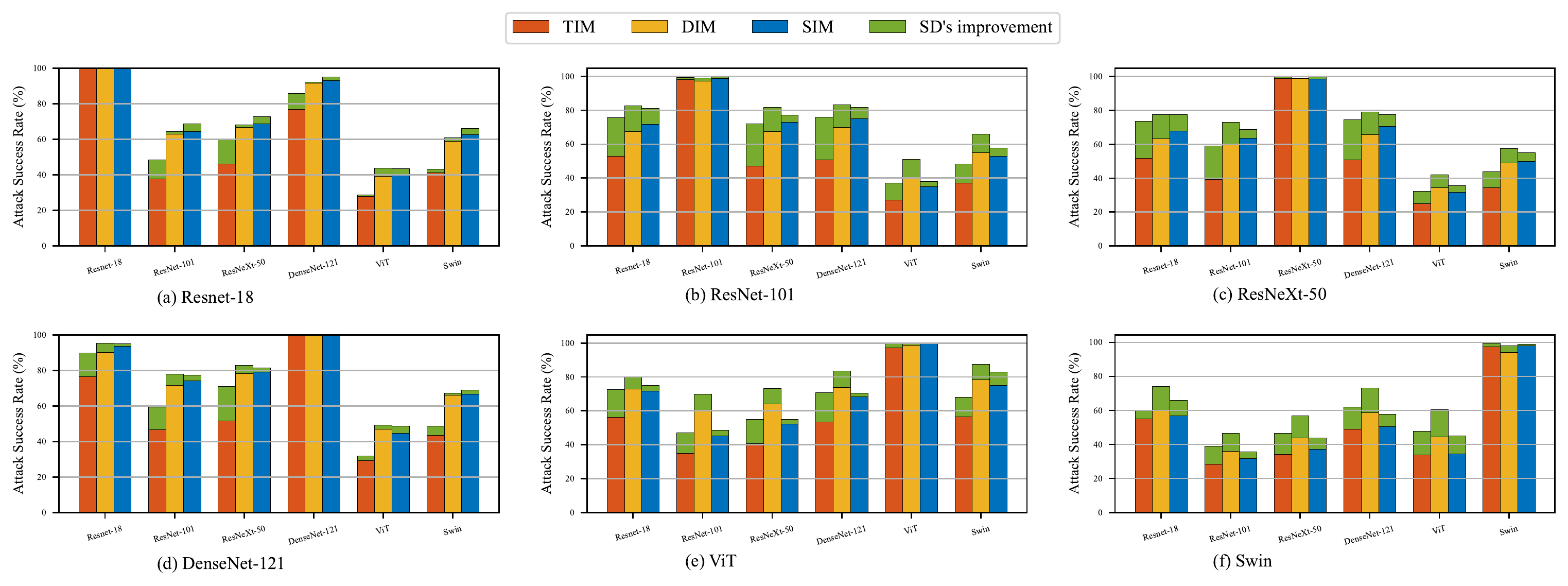}
    \caption{The Attack success rate (\%) of baseline models
on the crafted adversarial examples generated by DIM, TIM, SIM and our proposed method under single model setting. We generate the adversarial examples on one model and evaluate the transferability on the other five models. }
    \label{fig:input_performance}
\end{figure*}

\begin{figure}[htbp]
    \centering
    \includegraphics[scale=0.5]{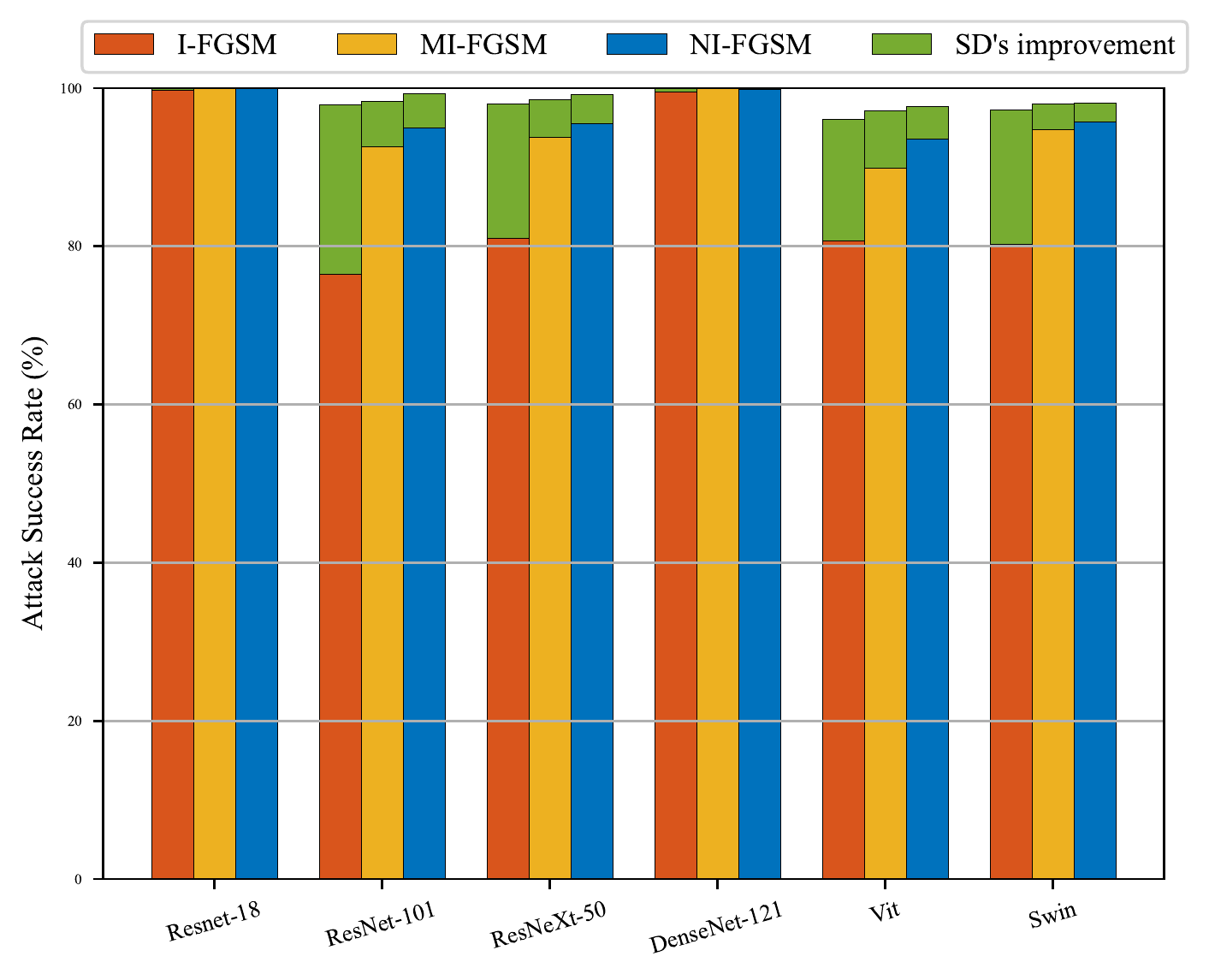}
    \caption{The Attack success rate (\%) of baseline models
on the crafted adversarial examples generated by I-FGSM, MI-FGSM, NI-FGSM and our proposed method under an ensemble model setting. We evaluate the performance on the hold-out model using the adversarial examples generated by the other five models.}
    \label{fig:ens}
    \end{figure}

In this section, we conduct extensive evaluations on ImaageNet dataset to validate the effectiveness of our proposed strategy, \textbf{S}chduled step size and \textbf{D}ual example (\textbf{SD}).

\subsection{Experimental Setup}
\noindent \textbf{Dataset}: We randomly select $1,000$ images remaining with $1,000$ categories from ILSVRC $2012$ validation set~\cite{krizhevsky2012imagenet}, which are almost classified correctly by the chosen models. 

\noindent \textbf{Victim Models}: We evaluate the attack performance on six popular convolution neural networks, including ResNet-18~\cite{he2016deep}, ResNet-101~\cite{he2016deep}, ResNeXt-50~\cite{xie2017aggregated}, DenseNet-121~\cite{huang2017densely}, Vision Transformer (ViT)~\cite{dosoViTskiy2020image} and Swin Transformer (Swin)~\cite{liu2021swin}. We also study several defense models, including the top-3 submissions in the NIPS 2017 defense challenge, namely High-level representation guided denoiser (HGD)~\cite{liao2018defense}, random resizing and padding (R\&P)~\cite{xie2017mitigating} and NIPS-r3~\footnote{\url{https://github.com/anlthms/nips-2017/tree/master/mmd}}, two adversarial training approaches, namely ensemble adversarially trained model (IncRes-v2$_{ens}$)~\cite{tramer2018ensemble} and Fast adversarial training (FastAdv)~\cite{wong2020fast}, one certified defense, namely randomized smoothing (RS)~\cite{cohen2019certified}, one deep denoiser, namely neural representation purifier (NRP)~\cite{naseer2020nrp}, and three input transformation-based defense method, namely JPEG~\cite{guo2017countering}, BitRed~\cite{xu2017feature}, and feature squeezing (FD)~\cite{liu2019feature}.

\noindent \textbf{Baselines}: For gradient-based attack methods, we adopt I-FGSM~\cite{kurakin2017adversarial}, MI-FGSM~\cite{dong2018boosting} and NI-FGSM~\cite{Lin2020Nesterov} as our baselines. For input transformation-based attacks, we treat DIM~\cite{xie2019improving}, TIM~\cite{dong2019evading} and SIM~\cite{Lin2020Nesterov} as our baselines.

\noindent \textbf{Hyper-parameters}: For the hyper-parameters, we follow the same setting of MI-FGSM~\cite{dong2018boosting} with the maximum perturbation of $\epsilon=16$, the number of iterations $T=10$, step size $\alpha=1.6$,  and decay factor $\mu=1.0$. Besides, we set the transformation probability for DIM as $0.5$~\cite{xie2019improving}, the size of Gaussian kernel for TIM as $7 \times 7$~\cite{dong2019evading}, and the number of scale copies for SIM as $m=5$~\cite{Lin2020Nesterov}. For our method, we set $\rho=0.9$ when we integrate the proposed strategy in MI-FGSM and NI-FGSM.    

\subsection{Integrated with Gradient-based Attacks}

\label{sec:exp:gradient}
Gradient-based adversarial attacks are the widely adopted approaches to craft adversarial examples. Here we first evaluate the effectiveness of \name to boost the attack performance of existing gradient-based attacks, including I-FGSM, MI-FGSM and NI-FGSM. The adversarial examples are generated on each model and evaluated on the other models. Te attack success rates, which are the misclassification rates of the victim models on the adversarial examples, are reported in Fig.~\ref{fig:baseline_performance}.

In general, we can observe that \name can effectively improve the attack performance of the baselines in white-box as well as black-box settings. On average, \name helps I-FGSM achieves the attack performance of $92.5\%$, which is $19.4\%$ higher than vanilla I-FGSM. Under the white-box setting, MI-FGSM and NI-FGSM achieves better transferability than I-FGSM since the momentum helps stablilize the optimization and escape local minima. Compared with these attacks, \name can significantly boost the transferability. In particular, \name can remarkably enhance the attack performance of I-FGSM, which outperforms NI-FGSM and MI-FGSM with a clear margin. \name can boost I-FGSM, NI-FGSM and MI-FGSM at least $15.7\%$, $6.7\%$, and $5.5\%$, respectively, showing its high effectiveness in improving the transferability and generality to various models.

\subsection{Integrated with Transformation-based Attacks}


Transformation-based adversarial attack  leverages multiple transformations on the inputs to improve the adversarial transferability. We are inline with the aforementioned experiments to integrate our method on three classical transformation-based attacks, namely DIM, TIM and SIM.  The reuslts can be shown in Fig. \ref{fig:input_performance}. 

We can observe that all the transformation-based methods  can achieve the attack success rate of $100.0\%$ or near $100.0\%$, showing that the transformation-based methods do not degrade and  even enhance the white-box attack performance. As for the black-box performance, TIM behaves the poorest transferability on these normally trained models.  Surprisingly, on some models, with \name strategy, TIM achieves even better transferability than DIM and even SIM. For instance, on ResNeXt-50, TIM with \name strategy achieves the attack success rate  of $79.9\%$, which is much high than $50.1\%$ of DIM and $70.8\%$ of SIM.  On average of all experiment cases, \name can boost the transferability of TIM, DIM and SIM with a clear margin at least  $4.9\%$, $2.1\%$ and $2.8\%$ respectively.

\begin{figure}[t]
    \centering
    \includegraphics[scale=0.5]{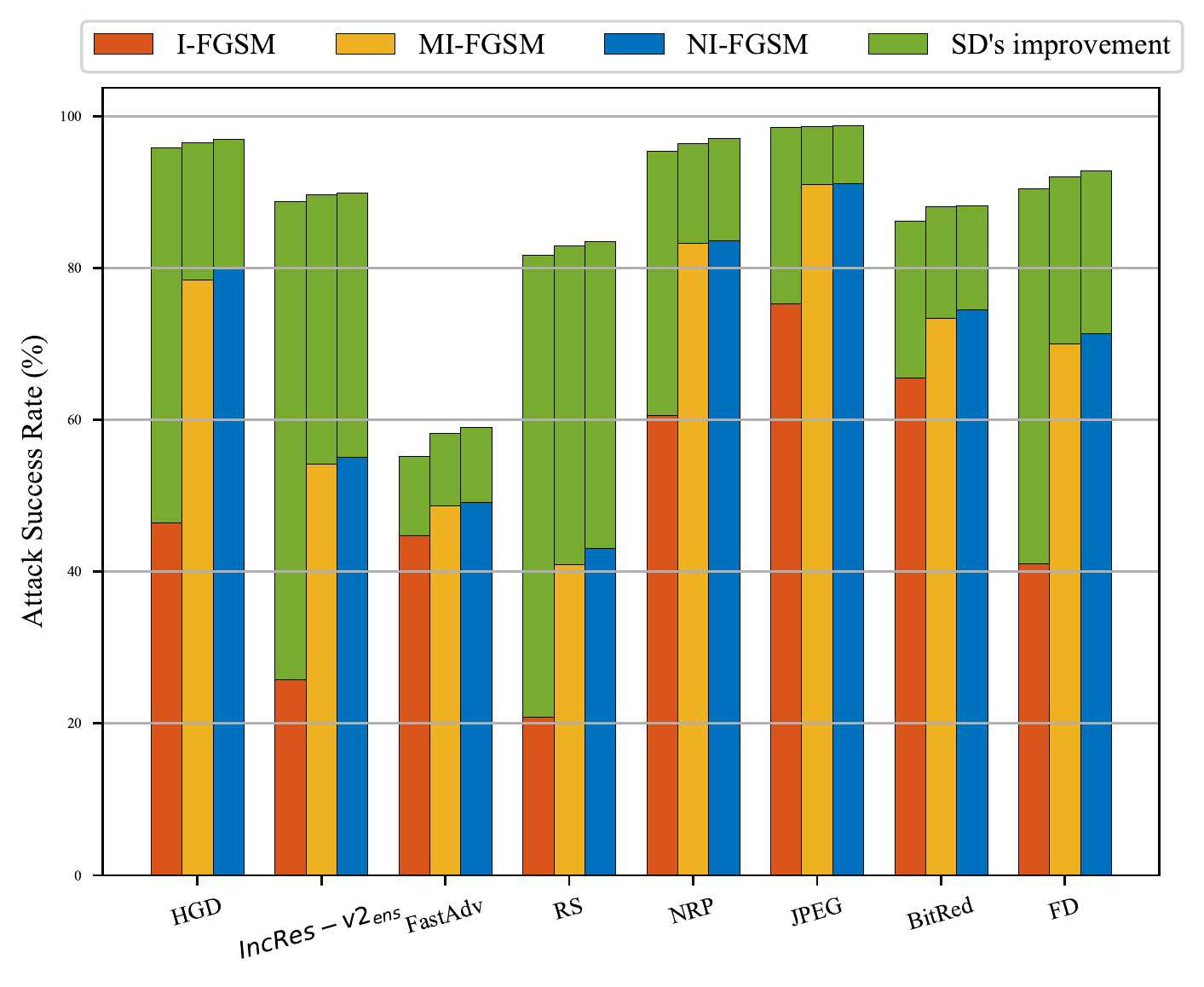}
    \caption{The Attack success rate (\%) of baseline models
on the crafted adversarial examples generated by I-FGSM, MI-FGSM, NI-FGSM and our proposed method under ensemble model setting. We evaluate on the advanced denfese methods. }
\label{fig:defense}
\end{figure} 

\begin{figure*}[htbp]
    \centering
    \includegraphics[width=\linewidth]{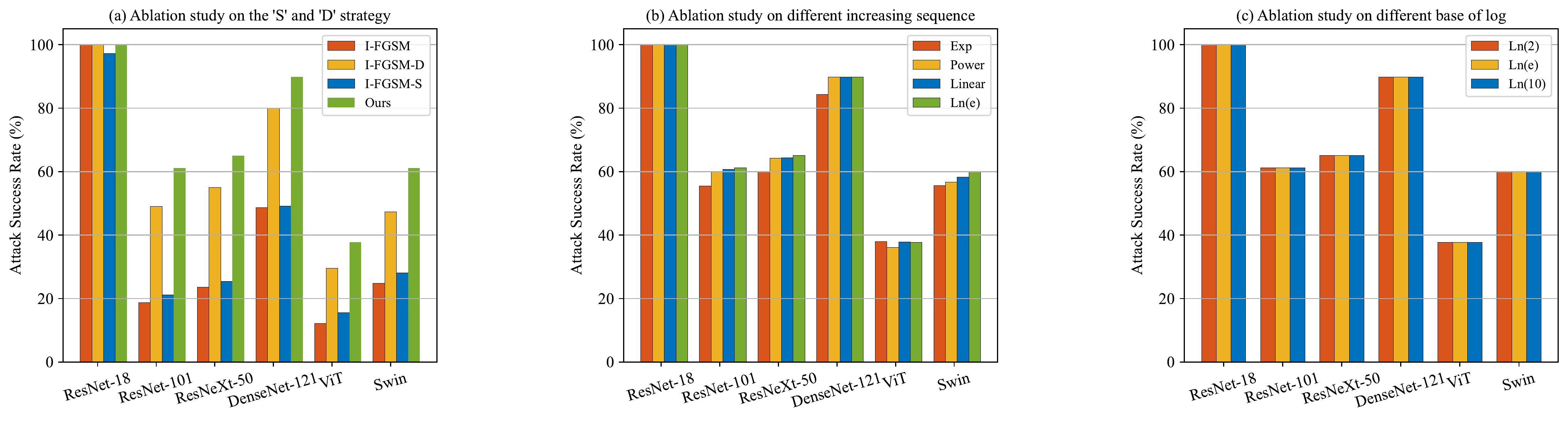}
    \caption{Ablation studies using the increasing sequence.}
    \label{fig:ablation}
    \end{figure*} 

\subsection{Integrated with Ensemble Attacks}

Ensemble attacks are another effective method to generate more transferable adversarial examples by attacking several models in parallel. We evaluate the performance of \name when integrated with the ensemble attack method. Specifically, we use the indicated single model as the evaluation model while we generate the adversarial example using the other five models. We present the results in Fig. \ref{fig:ens}. 

It can be clearly observed that \name strategy significantly enhances the adversarial attack performance, especially for I-FGSM. Specifically, with \name strategy can achieve an average attack success rate of $97.5\%$ , $98.4\%$,$99.2\%$ when integrated with I-FGSM, MI-FGSM and NI-FGSM, respectively, showing a clear margin of $16.2\%$, $3.0$ and $2.3\%$ compared with the vanilla ones.

\subsection{Attack on Advanced Defense Methods}
    
To further identify the effectiveness of our method, we evaluate the performance of the crafted adversarial examples on six defense methods, including HGD, $IncRes-v2_{ens}$, FastAdv, RS, NRP, JPEG, BitRed and FD. The target model for JPEG and NRP is VGG19 and the other methods adopts the official models provided in the corresponding works. We generate the adversarial examples on the ensemble model, \ie, ResNet-18, ResNet-101, ResNeXt-50, DenseNet-121. ViT and Swin. The results are summarized in Fig. \ref{fig:defense}.

From the figure, \name still exhibits an excellent attack performance towards various advanced defense methods.  Among them, the certificate robustness defense method, Random Smoothing (RS), is the most effective against adversarial attacks, where I-FGSM, MI-FGSM and NI-FGSM only achieves the attack success rates of $20.9\%$, $41.0\%$ and $43.1\%$ respectively. \name can remarkably enhance the attack performance with a clear improvement  of $50.8\%$, $32.9\%$ and $30.4\%$, respectively. On average, \name can improve the adversarial transferability of  I-FGSM, MI-FGSM and NI-FGSM with a clear margin of $38.9\%$, $20.3\%$, and $19.8\%$, respectively, showing the high effectiveness of our \name.

\subsection{Ablation Studies}
In this subsection, we mainly study the effectiveness of the proposed strategy and the choice of scheduled adaptive step sizes. We adopt ResNet-18 as the victim model to conduct the following experiments.

\noindent \textbf{The effectiveness of the proposed strategy.} There are two major components in our proposed method, namely scheduled adaptive step size and dual example. As carefully discussed in section \ref{sec:method}, the scheduled adaptive step size is used to explore more on the region where there exits more transferable adversarial examples while the dual example is used  to craft adversarial examples with high transferability. Here, we design an ablation study to identify the effectiveness of each component as follows,
1) I-FGSM: conventional iterative fast gradient sign method with the identity step size; 2) I-FGSM-D: apply the dual example to I-FGSM, where we update the dual example with perturbation smoothing; 3) I-FGSM-S: I-FGSM with scheduled adaptive step size; 4) Ours: I-FGSM with scheduled adaptive step size and the dual example. The results can be shown in Fig. \ref{fig:ablation} (a). We can identify the effectiveness of the dual example from the comparison between I-FGSM and I-FGSM-D. Although it sacrifices a little performance under the white-box setting with only the scheduled step size,  the transferability of crafted adversarial examples is significantly improved.


\noindent \textbf{The choice of scheduled adaptive step size.} As our analysis in Section \ref{sec:motivation}, the density of transferable adversarial examples decreases with increasing the distance to the benign example, so we adopt the increasing sequence as  the scheduled step size to fully utilize the perturbations with good transferability. We study four increasing sequences, exponentiation (Exp), power (Power), linear (Linear), logarithm sequence (Ln($\cdot$)) respectively. The  experiment results are shown in Fig. \ref{fig:ablation}
 From the result in Fig. \ref{fig:ablation}(b), it can be shown that with the decreasing of second-order, the attack performance of the increasing sequences display an increasing trend, \ie from Exp to Ln(e). We additionally adopts different number of base in logarithm function in Fig. \ref{fig:ablation} (c). It can be shown there exits minor difference between the settings with different bases.

\section{Conclusion}

In our work, we find an interesting phenomenon, that the FGSM and the I-FGSM with few iterations exhibit a better adversarial transferability than I-FGSM with more iterations. We hold an assumption for this  phenomenon, that the adversarial perturbations are more transferable near the benign sample. With increasing distance of the adversarial example relative to the benign sample, there is an increasing difference on the feature attention between different models, caused by the different architectures and distribution shifts. To alleviate this catastrophic performance drop, we propose a novel strategy, which uses the scheduled step size and dual example (SD), to fully utilize the transferable adversarial perturbation near the benign sample. Extensive experiments shows that SD can significantly boost the attack performance of existing adversarial attack methods, including the gradient-based attacks, transformation-basd attacks and ensemble attacks. Also, SD can remarkably enhance the adversarial transferability of existing methods towards various advanced defensive methods.

%

{\small
\bibliographystyle{ieee_fullname}
\bibliography{egbib}
}

\end{document}